%% file: main.tex
\documentclass[runningheads]{llncs}

\usepackage{accv}

\usepackage{accvabbrv}

\usepackage{graphicx}
\graphicspath{{figures/}}
\usepackage{booktabs}
\usepackage{multirow}

\usepackage{hyperref}

\usepackage{orcidlink}

\let\subparagraph\undefined
\usepackage{titlesec}

\titlespacing*{\section}{0pt}{8pt}{3pt}
\titlespacing*{\subsection}{0pt}{6pt}{2pt}
\titlespacing*{\subsubsection}{0pt}{4pt}{2pt}

\usepackage{siunitx}

\begin{document}

\title{Learning Human Sports Motion Feasibility via Quantile Regression on Tracking Data}
\titlerunning{Learning Sports Motion Feasibility via Quantile Regression}

\author{Marc Gutiérrez-Pérez \and Antonio Agudo}
\authorrunning{Gutiérrez-Pérez and Agudo}
\institute{Institut de Robòtica i Informàtica Industrial, CSIC-UPC, Barcelona, Spain \\ 
\email{\{marc.gutierrez.perez, antonio.agudo\}@upc.edu}}

\maketitle

\input{abstract}

\input{introduction}

\input{related_work}

\input{methodology}

\input{experiments}




\section{Conclusion}
\label{sec:conclusion}

We framed TTA as lower-quantile regression of locomotion time over physically normalized, heading-invariant geometry, augmented the pinball objective with a differentiable area-aware term, trained it on a $64$-match World~Cup tracking corpus under a strict cross-match split, and benchmarked it against six literature motion models re-expressed as arrival-time predictors and refit at a matching $0.99$ correctness gate, together with a learned reachable-hull baseline. The learned MLP beats every baseline on point error, RMSE and pinball loss throughout every distance band, and a single per-player max-speed feature improves on it further. At matched coverage, a quantile MLP is also the tighter one by reachable area in eight of the nine coverage$\times$horizon cells swept, with the profile variant taking every $95\%$ and $99\%$ cell at the longer $\Delta t \in \{3,5\}$\,s horizons. Retrained unchanged on basketball and American-football tracking, each sport's own model beats that sport's two-segment baseline on every metric in-domain, and cross-sport transfer is markedly asymmetric: the American-football-trained checkpoint transfers out essentially for free on point error and pinball, whereas every transferred checkpoint's reachable area loses to the destination's own physics baseline, a mismatch traced to a difference in label regime rather than to the representation. A fixed closed-form motion model is therefore not necessary to produce a physically plausible, well-calibrated reachable-area estimate. Natural next steps include richer conditioning for player personalization, and testing transfer across leagues and tracking providers within soccer itself.

\noindent \textbf{Acknowledgment.} This work has been supported by the project GRAVATAR PID2023-151184OB-I00 funded by MCIU/AEI/\allowbreak 10.13039/\allowbreak 501100011033 and by ERDF, UE and by the Government of Catalonia under Joan Oró FI 2024 grant.

%
%

\end{document}

%% file: abstract.tex
\begin{abstract}
We study \emph{time-to-arrival} (TTA) in sports: the shortest plausible time a player needs to reach an arbitrary location on the pitch given their current kinematic state, estimated as the lower envelope of observed locomotion times. Rather than modeling space ownership through position density estimation or closed-form physics-based models, we cast TTA as a conditional quantile regression problem, estimating the lower quantile of locomotion time so the model learns a player's lower-envelope arrival time. We train the model on optical tracking data from all $64$ matches of the PFF~FC FIFA World Cup~2022 dataset, and evaluate a lightweight multilayer perceptron against six player-motion models from the literature, re-expressed as TTA
predictors, and a learned reachable-hull baseline. Across our evaluation metrics, the MLP consistently outperforms every physics-based baseline. To probe the generality of the learned locomotion model, we further evaluate it on basketball tracking data and a single-game NFL tracking dataset, analyzing the transferability of player motion dynamics across sports. Our project is available at \url{https://github.com/mguti97/TTA-Quantile}.

\end{abstract}

%% file: introduction.tex
\section{Introduction}
\label{sec:intro}

Over the last decade, professional team sports have undergone a shift from box-score event logs to continuous positional tracking. Optical, radar, and GPS systems now record the locations of every player and the ball multiple times per second in sports such as football~\cite{somers2024soccernet, gutierrez2025soccernet}, basketball~\cite{sampaio2015exploring}, and American football~\cite{gudmundsson2017spatiotemporal}. This shift has opened a new class of technical, tactical and physical performance indicators~\cite{stein2017sense, gutierrez2026once}.

Space is the currency of team sports: where a player can be a second from now determines who can receive a pass, who can close a passing lane, and which regions of the pitch or court a team effectively controls. Almost every method that quantifies that (pitch control, off-ball value or pass-probability surfaces) rests, explicitly or implicitly, on a player-motion model: an answer to the question ``where can this player get to, and how fast?'' The dependence spans sports: recent models of expected passes~\cite{anzer2022expected}, dangerous
accessible space~\cite{bischofberger2026dangerous}, offside
control~\cite{novillo2025offside} and pressing
intensity~\cite{bekkers2025pressing} in football, off-ball value in basketball~\cite{wu2023obtracker}, and safe pass
targeting in American football~\cite{schilling2026safe} all inherit such a model, typically with a handful of fixed global parameters. Although developed in the context of sports analytics, player-motion models address a more general problem of predicting the future reachability of moving agents under physical constraints~\cite{Capellera_2024_ACCV,capellera2026jointdiff}. Similar formulations arise in robotics, where autonomous systems must anticipate the motion of humans or other robots for safe navigation and collision avoidance~\cite{unhelkar2018human, cai2023prediction}, and in autonomous driving for forecasting the reachable regions of surrounding vehicles and pedestrians~\cite{mahjourian2022occupancy, kim2022stopnet}.

Two answers to the previous question dominate the literature. One line derives a reachable region analytically from an equation of motion and then validates or refits that closed form against tracking~\cite{renkin2022validation,wu2025metric,
narizuka2023validation}. These models are interpretable and cheap, but they commit in advance to a fixed functional form with a handful of global parameters. A second line keeps a continuous notion of space but drops arrival time altogether, shaping a hand-set Gaussian influence kernel around a player's velocity and position~\cite{
fernandez2018wide,wu2023obtracker} or fitting a kernel-density estimate over observed endpoints~\cite{brefeld2019probabilistic,martens2021space,
spencer2019motion}. The kernel bandwidth may now be fit from data, but the output is a density or a behavioral commitment likelihood, not a feasibility bound: a location scores high because players are often there, or often choose it, independent of whether anyone could actually get there in time.

We take a complementary, data-driven view that keeps that question central. We directly predict the minimum time-to-arrival (TTA) from a player's kinematic state to any target location, and we learn it as a quantile of observed locomotion. The key modeling choice is to estimate a low envelope of arrival times rather than the mean: a player may dawdle to any location, so the upper tail is uninformative, but the floor on how quickly they can plausibly arrive is sharply defined and is exactly the quantity downstream space models need. We estimate this floor at $\tau=0.01$  rather than a looser quantile, since the target quantity is a feasibility bound (the player could be there in time). This turns an ill-posed counterfactual into a well-posed conditional-quantile regression problem. Unlike the iterative pursuit simulations and optimization-based fits some physical models require, the resulting predictor is a single compact feed-forward pass that runs comfortably beyond real time.

This work makes the following key contributions:
  \begin{itemize}
    \item \textbf{TTA as lower-quantile regression.} We cast time-to-arrival as pinball regression over physics-normalized, heading-invariant geometry.
    \item \textbf{An area-aware objective and a fair, joint benchmark.} We add a differentiable reachable-area term to the pinball loss and evaluate jointly on coverage and area against six literature motion models.
    \item \textbf{Cross-sport transferability analysis of a learned locomotion model.} We test the same representation and objective on basketball and American-football.
  \end{itemize}

%% file: related_work.tex
\section{Related work}
\label{sec:related}

\noindent\textbf{Physical motion models.}
The oldest strand models the region a player can reach in a given time from simple
kinematics. Taki and Hasegawa~\cite{taki2000visualization} introduced a
dominant-region idea using a constant-acceleration motion model, and Fujimura and
Sugihara~\cite{fujimura2005geometric} refined the locomotion law with a resistive-force model that caps speed smoothly. Efthimiou~\cite{efthimiou2022voronoi} generalizes the Voronoi partition to unequal characteristic speeds and reaction times, still a fixed closed-form law with hand-set per-player constants.  Caetano
{\em et al.}~\cite{caetano2021dominant} fit dominant regions from instantaneous
kinematic variables on real tracking. Renkin {\em et al.}~\cite{renkin2022validation}
provided the most systematic treatment we build on: they cataloged a family of motion models (constant speed, constant acceleration, capped acceleration, two-segment) and validated them against tracking by a reachable-area score, optimizing each model's parameters to maximize precision at a fixed correctness
gate. Wu and Swartz~\cite{wu2025metric} propose a free-direction constant-acceleration model with a speed cap and a corresponding pitch-control metric. Narizuka {\em et al.}~\cite{narizuka2023validation} validate the friction-damped model directly against J-League tracking, confirming circular rather than elliptical arrival regions.  Narizuka and Yamazaki~\cite{narizuka2025attacker} couple attacker and defender
  into the two-agent resistive-force pursuit of Brink {\em et
  al.}~\cite{brink2023dribbling}, which falls outside the single-player,
  fixed-target protocol we benchmark. Augustine {\em et al.}~\cite{augustine2025explainable} reuse the
  capped-acceleration model to time defenders in American football. We
adopt all six of these models as TTA predictors and fit them with exactly Renkin {\em et al.}'s reachable-area objective~\cite{renkin2022validation}, so our physics baselines are each model's best-tuned variant.\\

\noindent\textbf{Probabilistic and data-driven movement models.} A second strand drops the fixed functional form and estimates movement directly from data instead. Brefeld {\em et al.}~\cite{brefeld2019probabilistic} reframe
deterministic reachable regions as probabilistic movement models and zones of control estimated from data, and Lee and
Kitani~\cite{lee2016widereceiver} forecast a receiver's trajectory in American football with inverse-reinforcement learning over a data-driven opponent-motion
predictor, replacing kinematics with a learned, non-mechanistic trajectory model entirely. Both trade the interpretability of a closed-form law for flexibility fitted directly to tracking, the same trade-off the space-model literature below inherits, and the one our arrival-time formulation is designed to avoid. Anzer and Bauer~\cite{anzer2022expected} occupy an intermediate position we benchmark against directly: they estimate reachable regions directly from tracking data, as per-speed convex hulls of observed displacements, without committing to any
closed-form equation of motion. Several other data-driven movement models exist, but, Anzer and Bauer
aside, they estimate probabilistic occupancy rather than a physically interpretable space--time feasibility limit.

\noindent\textbf{Pitch control, space and pass models.}
A parallel strand turns motion models into team-level surfaces. Spearman {\em et al.}~\cite{spearman2017physics} build physics-based pass-probability surfaces from time-to-intercept, and Spearman~\cite{spearman2018beyond} extends this to
off-ball value. Fern\'andez and Bornn~\cite{fernandez2018wide} instead shape a heuristic Gaussian influence
kernel by velocity and ball distance, a continuous control surface with
expert-set parameters and no arrival-time semantics, and Caldeira
{\em et al.}~\cite{caldeira2024finishing} argue that static Voronoi space suffices in
finishing situations; both illustrate the alternative of discarding the time dimension altogether. The same pattern recurs outside soccer as Wu
{\em et al.}'s~\cite{wu2023obtracker} basketball visual-analytics system scores
off-ball value with a static Gaussian kernel around a player's current position masked by a Voronoi defensive term, again with no velocity, acceleration or arrival-time variable anywhere in the model. Fern\'andez {\em et al.}'s later expected-possession-value framework~\cite{fernandez2021framework} composes this same Gaussian influence surface with learned pass, drive and shot sub-models rather than replacing it, so its much larger downstream system still inherits the original heuristic
reachability layer unchanged. Spencer {\em et al.}~\cite{spencer2019motion} go further still in Australian-rules football, replacing kinematics with a purely empirical kernel-density estimate of where a player has historically
chosen to commit given the ball's travel time, a behavioral propensity model with no notion of physical reachability at all. Narizuka {\em et al.}~\cite{narizuka2021space} go the opposite way and weight every pitch location by team-level arrival-time margins, a formulation that is explicitly agnostic to the arrival-time model plugged in. Martens {\em et al.}~\cite{martens2021space} learn space and control from data and evaluate a distance-to-time regression, the closest prior work to our point-error view. Dick
{\em et al.}~\cite{dick2022availability} model availability (who can receive a pass) computationally. These methods predominantly output probabilities and are evaluated as classifiers; our contribution is orthogonal, supplying the underlying time as a calibrated quantile that such
surfaces can be rebuilt from. 

\noindent\textbf{Applications resting on arrival times.}
A growing applied literature consumes arrival times wholesale. Anzer and
Bauer~\cite{anzer2022expected} identify the intended receiver of a pass by intersecting a ball-flight model with data-driven reachable regions, and find
that arrival-time slack features dominate their pass-difficulty model; Chawla
{\em et al.}~\cite{chawla2017classification} rate pass quality from dominant-region
features; Bischofberger and Baca~\cite{bischofberger2026dangerous} drive
pass-interception rates by the player-versus-ball arrival-time differential; Schilling {\em et al.}~\cite{schilling2026safe} intersect receiver, defender and ball reachable regions to find safe pass targets in American football. In the same sport, Yurko {\em et al.}~\cite{yurko2020going} build a continuous-time play-valuation framework whose ball-carrier model consumes Voronoi-tessellation features but predicts expected yards gained rather than an arrival time, underscoring that the
motion-model gap we address is not specific to soccer. Allegre and
Vuillemot~\cite{allegre2020disputed} visualize disputed areas where arrival times nearly tie; and Novillo {\em et al.}~\cite{novillo2025offside} and da~Silva
{\em et al.}~\cite{dasilva2025stamina} build offside-control and stamina metrics on the standard reaction-plus-constant-speed heuristic. Pressing has become a
particularly active consumer of this layer: Bekkers~\cite{bekkers2025pressing} retargets the Spearman-style time-to-intercept formula from a fixed pitch
location to a moving player or ball to score pressing intensity, while Merckx
{\em et al.}~\cite{merckx2021pressing} plug an off-the-shelf pitch-control
implementation into gradient-boosted classifiers to value pressing
situations; both treat the arrival-time model purely as a black-box feature source. Link and Hoernig~\cite{link2017possession} similarly lean on
arrival-time reasoning at its simplest: they decide whether a player can still reach the ball within a fixed one-second horizon using a constant-velocity extrapolation. The underlying arrival time for these approaches is inherited with fixed global parameters, and is validated at best indirectly through downstream outcomes. Two of their findings directly motivate this work: da~Silva {\em et al.}~\cite{dasilva2025stamina} show pitch-control
surfaces are markedly sensitive to per-player top speed (matching our ablation, where a per-player max-speed feature is the largest single gain), and the horizon-dependent parameters of~\cite{narizuka2023validation} show a fixed
functional form cannot serve all horizons. A calibrated learned arrival time is a drop-in replacement for exactly this layer.

The two strands above trade off in opposite directions: motion models are interpretable and parameter-light but committed to a fixed functional form, while data-driven space models are flexible but discard the physically
meaningful time in favor of a classifier probability. This work sits at their intersection by proposing a learned model that keeps the interpretable output type
of the motion-model literature (an arrival time, invertible to a reachable area) while inheriting the accuracy of a data-driven regressor, evaluated against both literatures on their own terms.

%% file: methodology.tex
\section{Method}
\label{sec:methodology}

\begin{figure}[t!]
  \centering
  \includegraphics[width=1\linewidth]{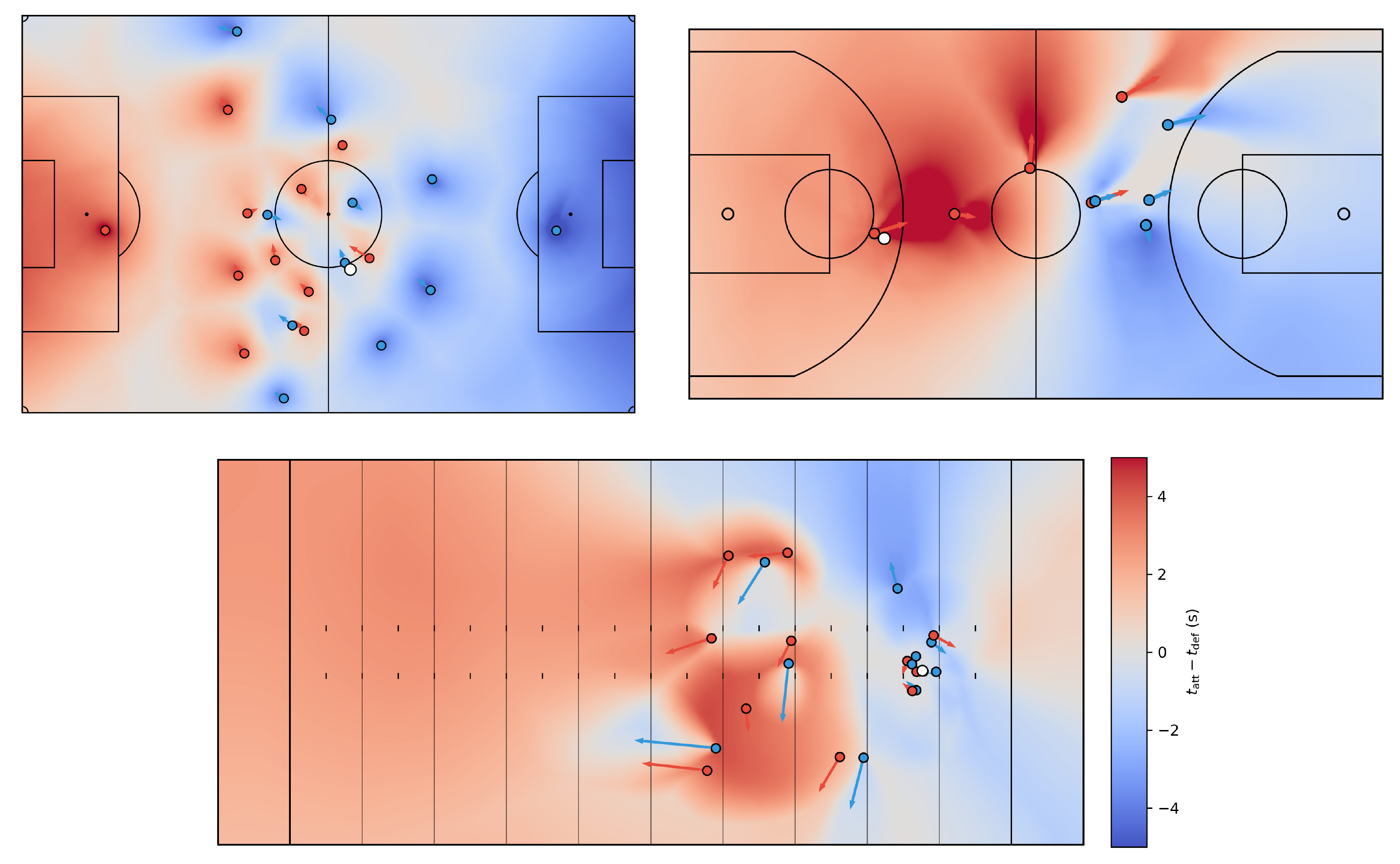}
\caption{\textbf{Predicted TTA surfaces for the attacking and defending teams} in football (top left), basketball (top right), and American football (bottom). The color map represents the difference between the predicted arrival times of the attacking and defending teams at each location. The ball is shown in white, and player arrows are scaled proportionally to player speed.}
  \label{fig:cross}
\end{figure}

We formulate TTA as conditional lower-quantile regression and present, in
  turn, the geometric duality that underpins both the model and its evaluation,
  the data pipeline from raw tracking to training tensors, the physical
  baselines re-expressed as arrival-time predictors, and the learned models and
  their objective. Figure~\ref{fig:cross} shows TTA surfaces in the three
  evaluated sports, as the difference in arrival time between attacking and
  defending team. 

\subsection{Problem formulation}
\label{sec:problem}

We predict the minimum time that a football player needs to
reach an arbitrary location on the pitch, given their current kinematic state $k = (\mathbf{p}, \mathbf{v}, \mathbf{a})$, where $\mathbf{p}$ is the position, $\mathbf{v}$ the velocity and
$\mathbf{a}$ the acceleration. Writing $\mathbf{x}$ for a target location, the quantity of interest is the arrival time $\Delta t(k \!\to\! \mathbf{x})$, which constitutes a counterfactual regression problem. At inference the model is queried
at locations the player never visited, so it must generalize from observed trajectories to hypothetical ones. The target is not where a player is expected to go but how fast they could plausibly get somewhere; that is, the lower envelope of observed locomotion. We therefore frame the task as conditional quantile regression~\cite{koenker1978regression}:
\begin{equation}
  \tau(\mathbf{g}) \;=\; Q_{0.01}\!\left(\Delta t \,\middle|\, \mathbf{g}(k, \mathbf{x})\right),
  \label{eq:quantile}
\end{equation}
where $\mathbf{g}(k,\mathbf{x})$ is the geometry of the state--target pair and $Q_{0.01}$ denotes the $1^{\text{st}}$ percentile of the arrival-time distribution. Estimating a low quantile, rather than the mean,
forces the model to hug the fastest reliably observed behavior while remaining
robust to the long upper tail of arrival times. We target the $1^{\text{st}}$ percentile, rather than a looser one, because the quantity of interest is a feasibility bound, not a typical-case estimate. The target is thus the \emph{conditional} lower envelope of arrival times, not a per-observation minimum. A single observed trajectory only upper-bounds what the player could have done; across the $\sim$154\,M extracted samples, however, many observations share the same kinematic state with the player fully committed to that displacement, and the pinball loss at the $1^{\text{st}}$ percentile recovers the fast tail of that conditional distribution, discarding choice-driven slowness rather than averaging it in. Intent, fatigue, tactical constraints and obstacles all act in the conservative direction.

\paragraph{Duality with reachable area.}
A TTA surface $\tau(\cdot)$ is the dual of a reachable-area model. The region a player can reach within a horizon $\Delta t$ is exactly the sublevel set:
\begin{equation}
  R(\Delta t) \;=\; \{\, \mathbf{x} : \tau(\mathbf{x}) \le \Delta t \,\}.
  \label{eq:sublevel}
\end{equation}
This identity connects our work to the player-motion-model literature~\cite{taki2000visualization,fujimura2005geometric,brefeld2019probabilistic,renkin2022validation,wu2025metric},
which models $R(\Delta t)$ directly, and it is the bridge that lets us evaluate a time output with a reachable-area metric.

\subsection{Track cleaning}
\label{sec:cleaning}

Each player's timeline within a match half is processed independently into a clean track of smoothed position, velocity and acceleration. Tracks are segmented into gap-free runs, smoothed with a Savitzky--Golay filter~\cite{savitzky1964smoothing} following the practice of~\cite{spearman2018beyond}, and differentiated by finite differences to obtain velocity and acceleration. Frames that violate hard kinematic bounds on speed, acceleration or frame-to-frame velocity change are excluded as unusable states. Exact window sizes and thresholds are provided in Supplementary Section~A.1.

\subsection{Coordinate transform and input features}
\label{sec:features}

Every state--target pair is expressed in the player's heading frame,
making the representation invariant to absolute pitch position and orientation. Given state position $\mathbf{p}$, velocity $\mathbf{v}$ (heading
$\varphi = \operatorname{atan2}(v_y, v_x)$, speed $v = \lVert\mathbf{v}\rVert$), acceleration $\mathbf{a}$, and a target $\mathbf{x}$, we (i)~translate so the
player sits at the origin and (ii)~rotate by $-\varphi$ so the heading aligns with $+x$. By construction the player moves along $+x$ at speed $v$, so the lateral velocity component vanishes and is dropped. The target then lands at
$(d\cos\theta,\, d\sin\theta)$, where $d$ is the distance to target and $\theta$ the angle from heading to target ($0^\circ$ ahead, $\pm180^\circ$ a full reversal). Acceleration is rotated into longitudinal/lateral components
$(a_x, a_y)$. The invariance deliberately discards absolute pitch position and boundary
context.

The network consumes six values, each normalized by a hard physical bound, so the same constants serve as clipping thresholds and feature scales (see Table~\ref{tab:features}). The angle is always encoded as the
pair $(\cos\theta, \sin\theta)$, never as a raw angle, which would introduce a $2\pi$ discontinuity.

\begin{table}[tb]
  \centering
\caption{The six input features and their normalization constants.}
  \begin{tabular}{lll}
    \toprule
    Feature & Meaning & Normalizer \\
    \midrule
    $d / 120$    & distance to target (m)            & $D_{\text{norm}} = 120$ \\
    $\cos\theta$ & cosine of angle to target         & $\in [-1, 1]$ \\
    $\sin\theta$ & sine of angle to target           & $\in [-1, 1]$ \\
    $v / 12$     & current speed (m/s)               & $V_{\text{norm}} = 12$ \\
    $a_x / 10$  & longitudinal acceleration (m/s$^2$) & $A_{\text{norm}} = 10$ \\
    $a_y / 10$  & lateral acceleration (m/s$^2$)    & $A_{\text{norm}} = 10$ \\
    \bottomrule
  \end{tabular}
  \label{tab:features}
\end{table}

\subsection{Sample extraction and label-quality filter}
\label{sec:samples}

For every usable state frame we emit training tuples
$(k, \mathbf{x}_{\Delta t}, \Delta t)$, where $\mathbf{x}_{\Delta t}$ indicates the
player's smoothed position $\Delta t$ seconds later. Each state draws five continuous horizon offsets uniformly from
$[0.10, 8.0]$\,s. Labels are therefore self-supervised: every target is the player's own observed future arrival time at the position they actually reached.

\noindent\textbf{SUVAT noise floor.}
SUVAT is a closed-form kinematic intercept model in the same family as the time-to-intercept models of~\cite{spearman2017physics,spearman2018beyond}. It runs in two phases in the heading frame: a redirect phase that keeps
the component of current velocity toward the target as useful momentum, followed by a sprint phase that accelerates at a capped rate up to a capped top speed and cruises.

Because the $\tau=0.01$ objective learns from the fastest realizations, a
single teleport artifact at a given geometry can drag the predicted floor below
what is physically achievable. We therefore use this model to compute, for every sample, the fastest physically plausible arrival time $\Delta t_{\text{SUVAT}}$ and discard any sample whose observed $\Delta t < \Delta t_{\text{SUVAT}}$, along with degenerate near-zero-distance samples. The filter is deliberately conservative: it removes only $0.3\%$ of
extracted samples, and its caps ($8$~m/s$^2$, $9$~m/s) are stricter than the independent
physical clipping bounds of Supplementary Section~A.1, so it discards
only samples already implausible under a model-free criterion. Retraining with the filter off, or with its caps widened to the model-free clipping bounds leaves the fitted floor unchanged and only degrades regularity: matched-coverage areas loosen by up to $31\%$ and $12$--$17\%$ respectively, and monotonicity violations
(Section~\ref{sec:structural-audit}) rise from $1.4\%$ to $23.4\%$ of rays. This is the signature of a noise screen, not of a physical prior, which would move the floor itself.

\subsection{Literature motion models as arrival-time predictors}
\label{sec:motion-models}

To benchmark against published player-motion models on a common footing, we re-express six models from the literature as TTA predictors and invert each via Eq.~\eqref{eq:sublevel}, $\tau(\mathbf{x}) = \min\{t : \mathbf{x} \in R(t)\}$. All operate in the heading frame, requiring only $(d, v, \cos\theta, \sin\theta)$; closed forms are provided in Supplementary Section~A.2.

\begin{itemize}
  \item \textbf{Constant speed} \cite{renkin2022validation}: a reachable disk that grows at a capped speed, centered on the player.
  \item \textbf{Constant acceleration} \cite{renkin2022validation,taki2000visualization}:
        a disk that grows quadratically and whose center drifts with the initial velocity. 
  \item \textbf{Constant acceleration with speed limit} \cite{renkin2022validation}:
        accelerates to a capped speed, then cruises; the boundary is no longer circular.
  \item \textbf{Two-segment constant speed} \cite{renkin2022validation}: an
        inertial drift segment followed by free radial motion.
  \item \textbf{Acceleration with free direction} \cite{wu2025metric}: a
        constant-magnitude acceleration whose direction is chosen freely on the circle, capped at a top speed, then cruising at constant velocity.
  \item \textbf{Friction-damped acceleration} \cite{fujimura2005geometric,brefeld2019probabilistic}:
        a driving force opposed by a resistive force proportional to speed, which caps speed smoothly rather than with a hard limit.
\end{itemize}
Each model's free parameters are fit to the training data by the same
reachable-area validation score used for evaluation,
so the comparison is between best-tuned versions of each model class.

\subsection{Learned models}
\label{sec:models}

\noindent\textbf{Multilayer perceptron (primary).}
The primary model is a small fully connected network,
$6 \to 64 \to 64 \to 1$, with ReLU hidden activations and a Softplus output. Because softplus of an affine map grows linearly, the network extrapolates gracefully beyond the training-horizon range.

\noindent\textbf{Player-profile MLP (personalized capacity).}
We augment the input with each player's maximum speed, estimated as the $99^{\text{th}}$ percentile of that player's cleaned, state-valid speeds pooled over the $45$ \emph{training} matches only. The scalar is normalized and appended as a seventh input.

\noindent\textbf{A learned data-driven baseline.} To separate the
\emph{quantile formulation} from mere data-drivenness, we additionally fit the
learned reachable-hull model of Anzer and Bauer~\cite{anzer2022expected},
which replaces the closed form with per-speed-bin convex hulls of the displacements observed in training tracking, evaluated under the identical protocol and on the identical splits.

\begin{figure}[t]
  \centering
  \includegraphics[width=1\linewidth]{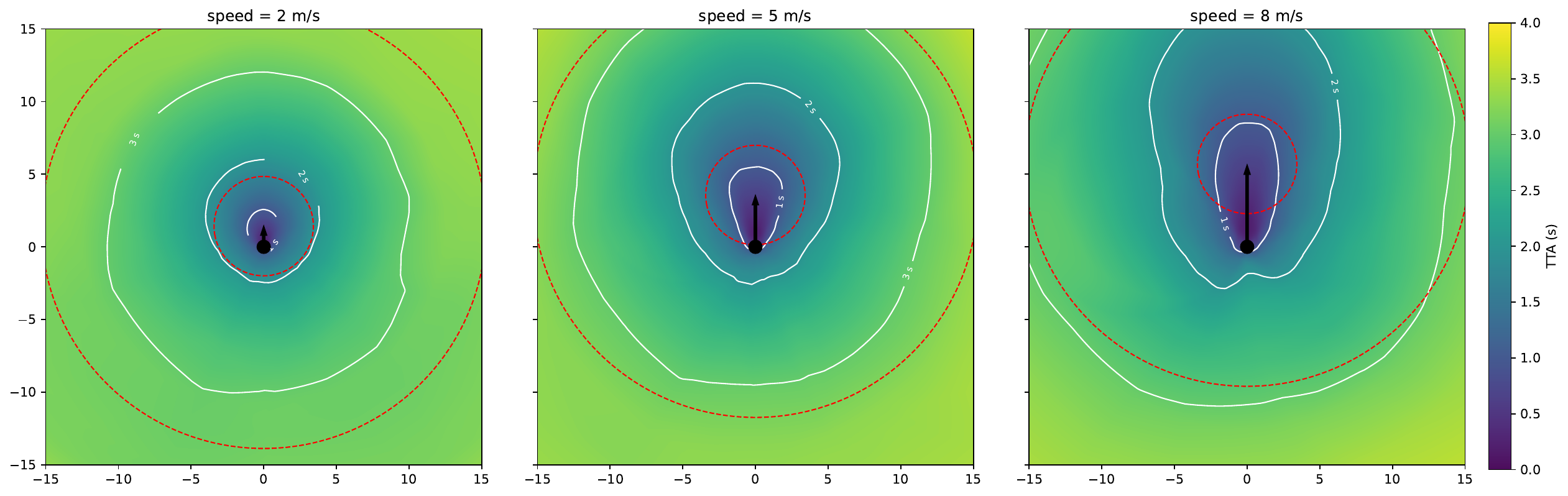}
  \caption{\textbf{Predicted TTA for a player moving upward at speeds of $2$, $5$, and $8$~m/s.} White contours correspond to the proposed learned model, while red ones correspond to the two-segment physical model~\cite{renkin2022validation}.}
  \label{fig:contours}
\end{figure}

Figure~\ref{fig:contours} depicts a visual comparison of time-contours for a player moving at speeds of 2, 5, and 8 m/s, with both physics-based and the proposed learned models.

\subsection{Quantile objective}
\label{sec:loss}

All learned models minimize a pinball loss based on~\cite{koenker1978regression}, augmented with an area-aware auxiliary term. The pinball component can be written as:
\begin{equation}
  \mathcal{L}_\tau(\hat{y}, y) =
  \begin{cases}
    \tau (y - \hat{y}) \;\;\;\;\;\;\;\;\;\;\;   \textrm{for}  & y \ge \hat{y} \\
    (1 - \tau)(\hat{y} - y) \;\;\textrm{for}  & y < \hat{y}
  \end{cases},
  \label{eq:pinball}
\end{equation}
where $y$ is the true observed TTA and $\hat{y}$ is the model's predicted arrival time for that same state--target pair. At $\tau = 0.01$, over-prediction is penalized $99\times$ more heavily than under-prediction, pushing the estimate toward the lower envelope so the model learns how quickly a player can arrive.

\noindent\textbf{Area-aware auxiliary term.}
  The pointwise pinball loss never sees the integrated pitch area that the learned surface produces through the duality of Eq.~\eqref{eq:sublevel}. We therefore add a differentiable relaxation of the reachable-area fraction. Because the surface is translation/rotation-invariant, its area can be evaluated on a fixed heading-frame grid $\mathcal{G}$ ($25\times25$ nodes at $4$\,m spacing spanning $\pm48$\,m, the origin node excluded, $|\mathcal{G}| = 624$); for a mini-batch subsample of states $k$ and horizons $h \in \{3, 5\}$\,s:
\begin{equation}
  \mathcal{L}_{\text{area}} \;=\;
  \operatorname*{mean}_{k,\,h}\;
  \frac{1}{|\mathcal{G}|} \sum_{\mathbf{x} \in \mathcal{G}}
  \sigma\!\left(\frac{h - \hat\tau_k(\mathbf{x})}{\beta}\right),
  \qquad \beta = 0.25\,\mathrm{s},
  \label{eq:arealoss}
\end{equation}
where $\sigma(\cdot)$ is the logistic sigmoid, a differentiable relaxation with $\beta$ setting the transition sharpness. The result is a soft count of the cells inside the horizon-$h$ reachable set. The total objective is $\mathcal{L} = \mathcal{L}_\tau + \lambda\,\mathcal{L}_{\text{area}}$, with
$\lambda = 0.01$.


%% file: experiments.tex
\section{Experiments}
\label{sec:experiments}

We now evaluate the proposed models, all of which are trained on identical samples, use the same six base features, and predict the same target quantile.
\subsection{Data}
\label{sec:data}

We use optical tracking data from the PFF~FC dataset of the FIFA World Cup 2022~\cite{PFFFC}, for all 64 matches, recorded at $30$\,fps, split by match at $45$/$10$/$9$ (train/validation/test). Holding out whole matches makes the 64-match corpus a genuine cross-match generalization test.

Tracking carries intermittent gaps (occlusions, substitutions), jitter that a
  double finite difference amplifies into implausible accelerations, and a small fraction of identity swaps or position jumps that sit precisely at the fast tail from which our $\tau=0.01$ objective learns (see Supplementary
  Section~A.1).

The six input features of Table~\ref{tab:features} are expressed entirely in physically normalized units (i.e., meters, m/s and m/s$^2$, respectively) within a player-centered heading frame. A checkpoint trained on one sport's samples can therefore be scored directly against another sport's held-out split, with no retraining or feature adaptation. To probe that, we build two further corpora through the identical pipeline of Sections~\ref{sec:cleaning}--\ref{sec:samples}.

For basketball, we use the NBA SportVU~\cite{NBAmov} player tracking, 2015--16 season, with $631$ games at $25$\,fps, split by match at $442$/$95$/$94$. Finally, for American football we consider the NFL Big Data Bowl 2019~\cite{BigDataBowl} tracking: a single game at $10$\,fps. Extraction yields 729,510 samples. Held-out splitting ($509$k/$87$k/$133$k) is by player rather than by match, a within-game rather than cross-match generalization test.

\subsection{Evaluation methodology}
\label{sec:eval}

To evaluate the point error, we report mean absolute error (MAE) and root-mean-square error (RMSE). We also evaluate the pinball loss of Eq.~\eqref{eq:pinball} at $\tau = 0.01$ on the held-out set.

\label{sec:eval-area}

Moreover, we port the player-motion-model validation idea of~\cite{renkin2022validation} through the duality of Eq.~\eqref{eq:sublevel}, but report it jointly as coverage and area. For a set of exact-$\Delta t$ trails reconstructed from the held-out split, with
$\mathbf{x}_t$ the point each player actually reached, coverage at horizon $\Delta t$ is the fraction of trails with $\tau_{\text{pred}}(\mathbf{x}_t)
\le \Delta t$ --- the reached point lies inside the predicted reachable set. The reachable area is the pitch area of the sublevel set $\{\mathbf{x} : \tau_{\text{pred}}(\mathbf{x}) \le \Delta t\}$, obtained by thresholding the inference grid.

Any model reaches coverage $1.0$ by inflating its reachable set to the whole pitch, so we never rank models on coverage or on area in isolation. Instead, we tune every model to a matched coverage and compare the resulting mean reachable area at that shared operating point: the model that covers the same fraction of actually-reached points with the smallest area is the tighter one. We do this across our five canonical horizons ($\Delta t \in \{1, 2, 3, 4, 5\}\,\mathrm{s}$) at the project's
$99\%$ coverage target.

\subsection{Main comparison}
\label{sec:main-results}

Our baselines are the physical arrival-time models: the six fitted literature
motion models of~\cite{renkin2022validation}, \cite{wu2025metric} and
\cite{fujimura2005geometric}, plus the learned reachable-hull model of Anzer and Bauer~\cite{anzer2022expected}. Table~\ref{tab:main} reports
point error, the pinball loss and matched-coverage reachable area for these seven motion models and the two quantile MLPs we propose. Both MLPs beat every physics baseline on MAE and pinball, by a wide margin. Moreover, appending a single per-player capacity feature is enough to improve on the base MLP: the profile variant lowers test point error and pinball, and tightens the matched-coverage area at the longer horizons, while matching it at $\Delta t=1$\,s, where acceleration out of the current state rather than top speed governs reach. The learned reachable-hull model trails both quantile MLPs on every
reported metric, and even the best closed-form baselines on point error: being data-driven is not, by itself, what produces a tight, well-calibrated lower envelope --- the
quantile formulation is.

\begin{table}[t!]
  \centering
  \caption{\textbf{Evaluation on PFF FC test-set.} Area is mean reachable area (m$^2$) at
  $\Delta t=1$\,s with every model tuned to the same $99\%$ coverage. Learned-model
  point metrics are mean$\pm$std over five training seeds.}
  \label{tab:main}
  \resizebox{\linewidth}{!}{%
  \begin{tabular}{llcccc}
    \toprule
    & Model & MAE (s) & RMSE (s) & pinball & area @\,99\%\,cov (m$^2$) \\
    \midrule
    \emph{physics}   & Constant speed \cite{renkin2022validation}        & 3.166 & 3.75 & 0.0334 & 119.3 \\
                     & Constant accel.\ \cite{renkin2022validation,taki2000visualization} & 3.043 & 3.63 & 0.0326 & 46.9 \\
                     & Const.\ accel.\ + $v_{\max}$ \cite{renkin2022validation} & 3.101 & 4.75 & 0.2370 & 48.0 \\
                     & Two-segment \cite{renkin2022validation}           & 2.992 & 3.60 & 0.0581 & \textbf{34.1} \\
                     & Two-segment $+\,v_{\max}$ (profile) \cite{renkin2022validation} & 2.900 & 3.50 & 0.0421 & 40.4 \\
                     & Free-direction accel.\ \cite{wu2025metric}        & \textbf{2.883} & \textbf{3.48} & 0.0307 & 43.6 \\
                     & Friction-damped \cite{fujimura2005geometric,brefeld2019probabilistic} & 2.984 & 3.57 & \textbf{0.0305} & 35.0 \\
    \midrule
    \emph{learned}   & Reachable hulls \cite{anzer2022expected} & 3.065 & 3.65 & 0.0308 & 34.8 \\
                     & MLP                               & $2.596_{\pm0.005}$ & $3.193_{\pm0.004}$ & $0.02800_{\pm0.00001}$ & \textbf{31.6} \\
                     & MLP $+\,v_{\max}$ (profile)        & $\textbf{2.577}_{\pm0.006}$ & $\textbf{3.171}_{\pm0.005}$ & $\textbf{0.02783}_{\pm0.00001}$ & 31.7 \\
    \bottomrule
  \end{tabular}%
  }
\end{table}

Table~\ref{tab:main}
reports each learned model's point metrics as mean\,$\pm$\,std over five training seeds, and the
profile model's advantage over the base model is positive in all five seeds. A paired cluster bootstrap over the
nine held-out test matches puts the base-versus-profile MAE gap at $+0.025$\,s with a $95\%$ confidence interval of $[+0.019, +0.030]$\,s, excluding zero. The gaps to every physics baseline are an order
of magnitude larger than this interval.

\begin{figure}[t]
  \centering
  \includegraphics[width=0.78\linewidth]{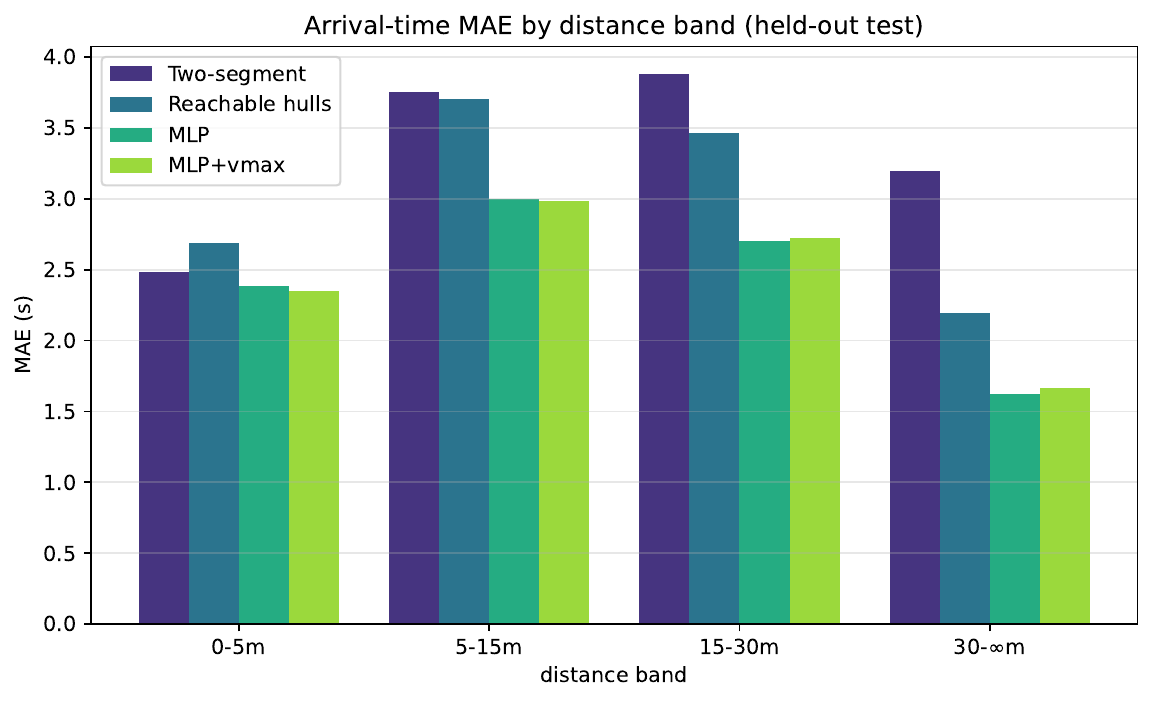}
  \caption{\textbf{MAE by distance band} on the PFF FC held-out test set, for the best precision physics baseline, the learned reachable-hull baseline and both quantile MLPs, stratified into four distance-to-target bands ($0$--$5$, $5$--$15$, $15$--$30$ and $30$\,m$+$).}
  \label{fig:stratified}
\end{figure}

Max speed allows the network to scale each player's reachable envelope individually rather than assuming one population-wide relationship between kinematic state and ceiling speed. Because
only the profile MLP receives per-player information, we ran a control in which the closed-form baselines are given the \emph{same} train-split per-player $v_{\max}$ table. The two-segment model benefits most, and we carry it through Table~\ref{tab:main} and Fig.~\ref{fig:area-sweep} as the representative personalized baseline. 
The pattern across the family is consistent: gains are small and concentrated at the long horizons, where a player's own ceiling speed actually binds. However, the best
personalized baseline still trails the profile MLP by $0.30$\,s of MAE. The advantage therefore stems from the
learned representation, not from the per-player feature.

In Fig.~\ref{fig:contours} is displayed a qualitative comparison of the predicted time contours produced by the physics-based and proposed learned models. Figure~\ref{fig:stratified} further breaks down the MAE by distance band, comparing the two-segment physical model and the learned reachable-hull baseline with both quantile MLPs. Across all distance bands, both MLPs outperform both baselines by a substantial margin.

\subsection{Ablation studies}
\label{sec:ablations}

\begin{figure}[t]
  \centering
  \includegraphics[width=\linewidth]{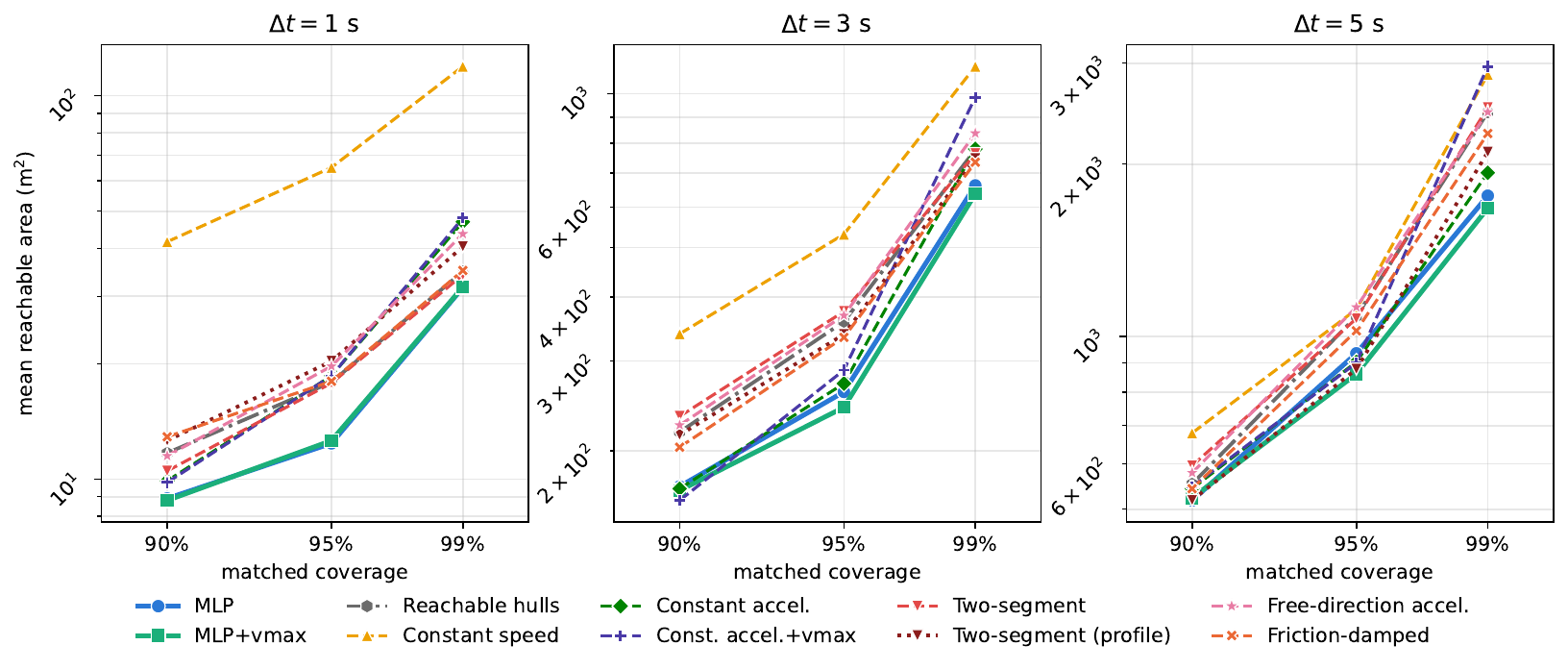}
  \caption{\textbf{Mean reachable area at matched coverage}, swept over three coverage levels ($90$/$95$/$99\%$) and three horizons ($\Delta t\in\{1,3,5\}$\,s), on PFF FC test-set.}
  \label{fig:area-sweep}
\end{figure}

\noindent\textbf{Reachable area and the coverage--precision trade-off.}
\label{sec:area-results}
Figure~\ref{fig:area-sweep} reports area at every combination of three coverage levels ($90$/$95$/$99\%$) and three horizons ($\Delta t\in\{1,3,5\}$\,s), for the two quantile MLPs, all six physics baselines and the learned reachable-hull baseline. A quantile MLP is tightest in eight of the nine cells. The learned reachable-hull baseline does not change the count: it is
looser than the base MLP in all nine cells, with the gap widening with horizon.

\label{sec:crosssport-results}
\noindent\textbf{Cross-sport transfer.} Table~\ref{tab:crosssport} shows the models' performance on the basketball and American-football corpora, each with its own $\tau=0.01$ MLP, then cross the three train/test checkpoint--split pairs to measure how well a locomotion model learned on one sport generalizes to another. 

In-domain, each sport's own MLP beats the two-segment baseline on every one of the four metrics, confirming the recipe is not tuned to one sport's geometry.

\begin{table}[t]
  \centering
  \footnotesize
  \caption{\textbf{Cross-sport transfer} for every sport-trained MLP checkpoint scored on each sport's own held-out test split, alongside that sport's own fitted two-segment baseline. NFL rows use within-game (player-based) splitting.}
  \label{tab:crosssport}
  \begin{tabular}{llcccc}
    \toprule
    Test & Train & MAE (s) & RMSE (s) & pinball & area @\,99\%\,cov (m$^2$) \\
    \midrule
    \multirow{4}{*}{Football}
      & Football    & \textbf{2.602} & \textbf{3.20} & \textbf{0.0280} & \textbf{31.6} \\
      & Basketball  & 2.631 & 3.36 & 0.0808 & 35.1 \\
      & NFL         & 2.652 & 3.25 & 0.0302 & 46.8 \\
      & Two-segment & 2.996 & 3.61 & 0.0581 & 35.7 \\
    \midrule
    \multirow{4}{*}{Basketball}
      & Basketball  & \textbf{2.462} & \textbf{3.08} & \textbf{0.0265} & \textbf{47.7} \\
      & Football    & 2.490 & 3.09 & 0.0276 & 73.0 \\
      & NFL         & 2.528 & 3.13 & 0.0280 & 59.9 \\
      & Two-segment & 2.945 & 3.56 & 0.0614 & 54.9 \\
    \midrule
    \multirow{4}{*}{Am.\ football}
      & NFL         & 1.297 & 1.82 & \textbf{0.0135} & \textbf{14.2} \\
      & Football    & \textbf{1.273} & \textbf{1.81} & 0.0165 & 28.3 \\
      & Basketball  & 1.284 & 2.17 & 0.1525 & 36.5 \\
      & Two-segment & 1.608 & 2.15 & 0.0891 & 20.0 \\
    \bottomrule
  \end{tabular}
\end{table}

Cross-sport transfer is markedly asymmetric. On point error and pinball, the American-football-trained checkpoint transfers out best: it beats the destination's own two-segment baseline in both directions, and its pinball stays within a few percent of the destination's own in-domain MLP. The reverse direction is weaker and
sport-dependent: football transfers into American football almost as well as
NFL's own model but basketball's slower, more lateral game sits furthest in
kinematic regime.

Every out-of-domain checkpoint in Table~\ref{tab:crosssport} has a larger area than the destination's own two-segment baseline, even where its MAE and pinball beat it: the lower envelope carries its point-error competitiveness across sports while producing a looser reachable set than a physics model tuned to the destination's own geometry. The mechanism is a mismatch in label regime rather than a failure of the representation: American-football pass plays are short, explosive, straight-line sprints, with faster arrival times at matched distance and speed than continuous-possession football or basketball play.

\noindent\textbf{Structural audit of the learned surface.}
\label{sec:structural-audit}
Nothing in the architecture enforces the properties a reachable-set
interpretation requires, so we audit them directly (protocol in Supplementary Section~A.4). On $5{,}000$ held-out
states the predicted sublevel sets are a single connected region containing the
player in $100\%$ of the $25{,}000$ state$\times$horizon cases examined, and are
star-shaped in $100\%$ of $80{,}000$ audited rays at every canonical horizon.
Radial monotonicity is violated on $1.4\%$ of rays, all of them at $\tau>4$\,s
and beyond $20$\,m from the player, with a median
dip of $16$\,ms --- far outside the operating range of the reachable-area
metric. The polygon representation the downstream literature assumes therefore
remains valid in practice.

\subsection{Real-time inference}
\label{sec:inference}

Both learned models are small enough to run online during a live match. The cost that matters in practice is a full reachable-area or pitch-control map: every player's surface evaluated over the $1$\,m heading-frame grid of the $105\times68$\,m pitch ($7{,}140$ cells), batched across all $22$ on-field players into one forward pass of $157{,}080$ rows. On an NVIDIA RTX~A5000 GPU this pass takes $2.4$\,ms, i.e.,\ $\sim\!420$ frames per second, an order of magnitude above the $25$--$30$\,fps a tracking system samples at.

%% file: main.bbl
\begin{thebibliography}{10}
\providecommand{\url}[1]{\texttt{#1}}
\providecommand{\urlprefix}{URL }
\providecommand{\doi}[1]{https://doi.org/#1}

\bibitem{allegre2020disputed}
Allegre, J., Vuillemot, R.: Visualizing and analyzing disputed areas in soccer. In: 2020 Visualization in Data Science (VDS). pp. 52--60. IEEE (2020)

\bibitem{anzer2022expected}
Anzer, G., Bauer, P.: Expected passes: Determining the difficulty of a pass in football (soccer) using spatio-temporal data. Data mining and knowledge discovery  \textbf{36}(1),  295--317 (2022)

\bibitem{augustine2025explainable}
Augustine, A., Redding, G.P., Moan, S.L.: Toward explainable data and sports analytics: A case study on pass completion prediction in american football. The American Statistician  \textbf{79}(4),  435--448 (2025)

\bibitem{bekkers2025pressing}
Bekkers, J.: Pressing intensity: An intuitive measure for pressing in soccer. arXiv preprint arXiv:2501.04712  (2024)

\bibitem{bischofberger2026dangerous}
Bischofberger, J., Baca, A.: Dangerous accessible space: a unified model of space and value in team sports. Journal of Big Data  (2026)

\bibitem{brefeld2019probabilistic}
Brefeld, U., Lasek, J., Mair, S.: Probabilistic movement models and zones of control. Machine Learning  \textbf{108}(1),  127--147 (2019)

\bibitem{brink2023dribbling}
Brink, L., Ha, S.K., Snowdon, J., Vidal-Codina, F., Rauch, B., Wang, F., Wu, D., L{\'o}pez-Felip, M.A., Clanet, C., Hosoi, A.E.: Measuring skill via player dynamics in football dribbling. Scientific Reports  \textbf{13}(1),  19004 (2023)

\bibitem{caetano2021dominant}
Caetano, F.G., Barbon~Junior, S., Torres, R.d.S., Cunha, S.A., Ruffino, P.R.C., Martins, L.E.B., Moura, F.A.: Football player dominant region determined by a novel model based on instantaneous kinematics variables. Scientific reports  \textbf{11}(1),  18209 (2021)

\bibitem{cai2023prediction}
Cai, J., Du, A., Liang, X., Li, S.: Prediction-based path planning for safe and efficient human--robot collaboration in construction via deep reinforcement learning. Journal of computing in civil engineering  \textbf{37}(1),  04022046 (2023)

\bibitem{caldeira2024finishing}
Caldeira, N., Lopes, R.J., Araujo, D., Fernandes, D.: The finishing space value for shooting decision-making in high-performance football. Sports  \textbf{12}(8), ~208 (2024)

\bibitem{Capellera_2024_ACCV}
Capellera, G., Ferraz, L., Rubio, A., Agudo, A., Moreno-Noguer, F.: Transportmer: A holistic approach to trajectory understanding in multi-agent sports. In: Proceedings of the Asian Conference on Computer Vision (ACCV) (2024)

\bibitem{capellera2026jointdiff}
Capellera, G., Ferraz, L., Rubio, A., Alahi, A., Agudo, A.: {JointDiff}: Bridging continuous and discrete in multi-agent trajectory generation. In: International Conference on Learning Representations (2026)

\bibitem{chawla2017classification}
Chawla, S., Estephan, J., Gudmundsson, J., Horton, M.: Classification of passes in football matches using spatiotemporal data. ACM Transactions on Spatial Algorithms and Systems (TSAS)  \textbf{3}(2),  1--30 (2017)

\bibitem{dick2022availability}
Dick, U., Link, D., Brefeld, U.: Who can receive the pass? a computational model for quantifying availability in soccer. Data mining and knowledge discovery  \textbf{36}(3),  987--1014 (2022)

\bibitem{efthimiou2022voronoi}
Efthimiou, C.J.: The voronoi diagram in soccer revisited. In: Future of Information and Communication Conference. pp. 500--525. Springer (2023)

\bibitem{fernandez2018wide}
Fernandez, J., Bornn, L.: Wide open spaces: A statistical technique for measuring space creation in professional soccer. In: Sloan sports analytics conference. vol.~2018 (2018)

\bibitem{fernandez2021framework}
Fern{\'a}ndez, J., Bornn, L., Cervone, D.: A framework for the fine-grained evaluation of the instantaneous expected value of soccer possessions. Machine Learning  \textbf{110}(6),  1389--1427 (2021)

\bibitem{fujimura2005geometric}
Fujimura, A., Sugihara, K.: Geometric analysis and quantitative evaluation of sport teamwork. Systems and Computers in Japan  \textbf{36}(6),  49--58 (2005)

\bibitem{gudmundsson2017spatiotemporal}
Gudmundsson, J., Horton, M.: Spatio-temporal analysis of team sports. ACM Computing Surveys (CSUR)  \textbf{50}(2),  1--34 (2017)

\bibitem{gutierrez2025soccernet}
Guti{\'e}rrez-P{\'e}rez, M., Agudo, A.: Soccernet-v{3D}: Leveraging sports broadcast replays for {3D} scene understanding. In: Proceedings of the Computer Vision and Pattern Recognition Conference. pp. 5978--5987 (2025)

\bibitem{gutierrez2026once}
Guti{\'e}rrez-P{\'e}rez, M., Yeung, C., Fujii, K., Agudo, A.: Once upon a goal: Towards orientation-based shot metrics in football. Computer Vision and Image Understanding p. 104732 (2026)

\bibitem{kim2022stopnet}
Kim, J., Mahjourian, R., Ettinger, S., Bansal, M., White, B., Sapp, B., Anguelov, D.: Stopnet: Scalable trajectory and occupancy prediction for urban autonomous driving. In: 2022 International Conference on Robotics and Automation (ICRA). pp. 8957--8963. IEEE (2022)

\bibitem{koenker1978regression}
Koenker, R., Bassett, G.: Regression quantiles. Econometrica: Journal of the Econometric Society  \textbf{46}(1),  33--50 (1978)

\bibitem{BigDataBowl}
League, N.F.: Nfl big data bowl 2019 (2019), \url{https://github.com/nfl-football-ops/Big-Data-Bowl}

\bibitem{lee2016widereceiver}
Lee, N., Kitani, K.M.: Predicting wide receiver trajectories in american football. In: 2016 IEEE Winter Conference on Applications of Computer Vision (WACV). pp.~1--9. IEEE (2016)

\bibitem{link2017possession}
Link, D., Hoernig, M.: Individual ball possession in soccer. PloS one  \textbf{12}(7),  e0179953 (2017)

\bibitem{mahjourian2022occupancy}
Mahjourian, R., Kim, J., Chai, Y., Tan, M., Sapp, B., Anguelov, D.: Occupancy flow fields for motion forecasting in autonomous driving. IEEE Robotics and Automation Letters  \textbf{7}(2),  5639--5646 (2022)

\bibitem{martens2021space}
Martens, F., Dick, U., Brefeld, U.: Space and control in soccer. Frontiers in Sports and Active Living  \textbf{3},  676179 (2021)

\bibitem{merckx2021pressing}
Merckx, S., Robberechts, P., Euvrard, Y., Davis, J.: Measuring the effectiveness of pressing in soccer. In: Proceedings of the 8th Workshop on Machine Learning and Data Mining for Sports Analytics (2021)

\bibitem{narizuka2023validation}
Narizuka, T., Takizawa, K., Yamazaki, Y.: Validation of a motion model for soccer players' sprint by means of tracking data. Scientific reports  \textbf{13}(1), ~865 (2023)

\bibitem{narizuka2025attacker}
Narizuka, T., Yamazaki, I.: Evaluating soccer player movements using the attacker-defender model. arXiv preprint arXiv:2512.22596  (2025)

\bibitem{narizuka2021space}
Narizuka, T., Yamazaki, Y., Takizawa, K.: Space evaluation in football games via field weighting based on tracking data. Scientific reports  \textbf{11}(1), ~5509 (2021)

\bibitem{novillo2025offside}
Novillo, {\'A}., del Campo, R.L., Serra, R.R., Buld{\'u}, J.M.: Offside control: A pitch control parameter to evaluate the offside performance of soccer teams. Chaos, Solitons \& Fractals  \textbf{197},  116445 (2025)

\bibitem{PFFFC}
{PFF FC}: {PFF FC} 2022 world cup dataset (2022), \url{https://www.blog.fc.pff.com/blog/pff-fc-release-2022-world-cup-data}

\bibitem{renkin2022validation}
Renkin, M., Bischofberger, J., Schikuta, E., Baca, A.: Validation and optimisation of player motion models in football. In: International Conference on Computational Science. pp. 26--32. Springer (2022)

\bibitem{sampaio2015exploring}
Sampaio, J., McGarry, T., Calleja-Gonz{\'a}lez, J., Jim{\'e}nez~S{\'a}iz, S., Schelling i~del Alc{\'a}zar, X., Balciunas, M.: Exploring game performance in the national basketball association using player tracking data. PloS one  \textbf{10}(7),  e0132894 (2015)

\bibitem{savitzky1964smoothing}
Savitzky, A., Golay, M.J.E.: Smoothing and differentiation of data by simplified least squares procedures. Analytical Chemistry  \textbf{36}(8),  1627--1639 (1964)

\bibitem{schilling2026safe}
Schilling, M., Milani, R., Moll, M., Pickl, S.: Modeling safe passing regions in american football: integrating ball trajectories and player coverage. Journal of Big Data  (2026)

\bibitem{dasilva2025stamina}
da~Silva, M.F., Novillo, A., Aleta, A., del Campo, R.L., Resta, R., Moreno, Y., Buldu, J.: Interplay between pitch control and top speed in soccer: The stamina factor. Scientific Reports  \textbf{15}(1),  38799 (2025)

\bibitem{somers2024soccernet}
Somers, V., Joos, V., Cioppa, A., Giancola, S., Ghasemzadeh, S.A., Magera, F., Standaert, B., Mansourian, A.M., Zhou, X., Kasaei, S., et~al.: Soccernet game state reconstruction: End-to-end athlete tracking and identification on a minimap. In: Proceedings of the IEEE/CVF Conference on Computer Vision and Pattern Recognition. pp. 3293--3305 (2024)

\bibitem{spearman2018beyond}
Spearman, W.: Beyond expected goals. In: Proceedings of the 12th MIT sloan sports analytics conference. pp. 1--17 (2018)

\bibitem{spearman2017physics}
Spearman, W., Basye, A., Dick, G., Hotovy, R., Pop, P.: Physics-based modeling of pass probabilities in soccer. In: Proceeding of the 11th MIT Sloan Sports Analytics Conference. vol.~1. Boston, MA (2017)

\bibitem{spencer2019motion}
Spencer, B., Jackson, K., Robertson, S.: Fitting motion models to contextual player behavior. In: International symposium on computer science in sport. pp. 11--18. Springer (2019)

\bibitem{NBAmov}
SportVU: {NBA} player movements (2017), \url{https://github.com/linouk23/NBA-Player-Movements}

\bibitem{stein2017sense}
Stein, M., Janetzko, H., Seebacher, D., J{\"a}ger, A., Nagel, M., H{\"o}lsch, J., Kosub, S., Schreck, T., Keim, D.A., Grossniklaus, M.: How to make sense of team sport data: From acquisition to data modeling and research aspects. Data  \textbf{2}(1), ~2 (2017)

\bibitem{taki2000visualization}
Taki, T., Hasegawa, J.: Visualization of dominant region in team games and its application to teamwork analysis. In: Proceedings computer graphics international 2000. pp. 227--235. IEEE (2000)

\bibitem{unhelkar2018human}
Unhelkar, V.V., Lasota, P.A., Tyroller, Q., Buhai, R.D., Marceau, L., Deml, B., Shah, J.A.: Human-aware robotic assistant for collaborative assembly: Integrating human motion prediction with planning in time. IEEE Robotics and Automation Letters  \textbf{3}(3),  2394--2401 (2018)

\bibitem{wu2025metric}
Wu, L., Swartz, T.B.: A new metric for pitch control based on an intuitive motion model. Computational Statistics  \textbf{40}(4),  1713--1730 (2025)

\bibitem{wu2023obtracker}
Wu, Y., Deng, D., Xie, X., He, M., Xu, J., Zhang, H., Zhang, H., Wu, Y.: Obtracker: Visual analytics of off-ball movements in basketball. IEEE Transactions on Visualization and Computer Graphics  \textbf{29}(1),  929--939 (2022)

\bibitem{yurko2020going}
Yurko, R., Matano, F., Richardson, L.F., Granered, N., Pospisil, T., Pelechrinis, K., Ventura, S.L.: Going deep: Models for continuous-time within-play valuation of game outcomes in american football with tracking data. Journal of Quantitative Analysis in Sports  \textbf{16}(2),  163--182 (2020)

\end{thebibliography}
